%% file: arxiv.tex
\pdfoutput=1

\documentclass[11pt]{article}

\usepackage[final]{acl}

\usepackage{times}
\usepackage{latexsym}

\usepackage[T1]{fontenc}

\usepackage[utf8]{inputenc}

\usepackage{microtype}

\usepackage{inconsolata}

\usepackage{graphicx}

\usepackage{booktabs}
\usepackage{colortbl}
\usepackage{multirow}
\usepackage{xspace}

\newcommand{\revise}[1]{\textcolor{black}{#1}}

\input{math_commands.tex}

\title{PSLM: Parallel Generation of Text and Speech with LLMs\\for Low-Latency Spoken Dialogue Systems}

\author{
Kentaro Mitsui, Koh Mitsuda, Toshiaki Wakatsuki, Yukiya Hono, Kei Sawada \\ \\
  rinna Co., Ltd., Tokyo, Japan \\
  \texttt{\{kemits,kohmi,towaka,yuhono,keisawada\}@rinna.co.jp}\\
}
\begin{document}

\renewcommand{\arraystretch}{0.8}
\setlength{\abovedisplayskip}{5pt}
\setlength{\belowdisplayskip}{5pt}

\maketitle
\begin{abstract}
Multimodal language models that process both text and speech have a potential for applications in spoken dialogue systems.
However, current models face two major challenges in response generation latency: 
(1) generating a spoken response requires the prior generation of a written response, and
(2) speech sequences are significantly longer than text sequences.
This study addresses these issues by extending the input and output sequences of the language model to support the parallel generation of text and speech.
Our experiments on spoken question answering tasks demonstrate that our approach improves latency while maintaining the quality of response content.
Additionally, we show that latency can be further reduced by generating speech in multiple sequences.
Demo samples are available at \url{https://rinnakk.github.io/research/publications/PSLM}.
\end{abstract}

\section{Introduction}

Spoken dialogue systems have been developed for many years to achieve natural human-computer interaction~\citep{mctear2002spoken,jokinen2009spoken,hongshen2017survey}.
Traditionally, these systems consist of several components: Automatic Speech Recognition (ASR), Response Generation (RG), and Text-to-Speech (TTS).
Various methods for RG have been proposed with the advancements in Large Language Models (LLMs)~\citep{wang2023survey,yi2024survey}.
More recently, the application of LLMs to ASR~(e.g., \citealt{wang2023viola,hono-etal-2024-integrating,fathullah2024prompting}) and TTS~\citep{wang2023viola,hao2023boosting} has attracted much attention, leading to the development of multimodal LLMs capable of end-to-end spoken language communication~\citep{zhang-etal-2023-speechgpt,nachmani2024spoken}.

\begin{figure}[t]
  \includegraphics[width=\columnwidth]{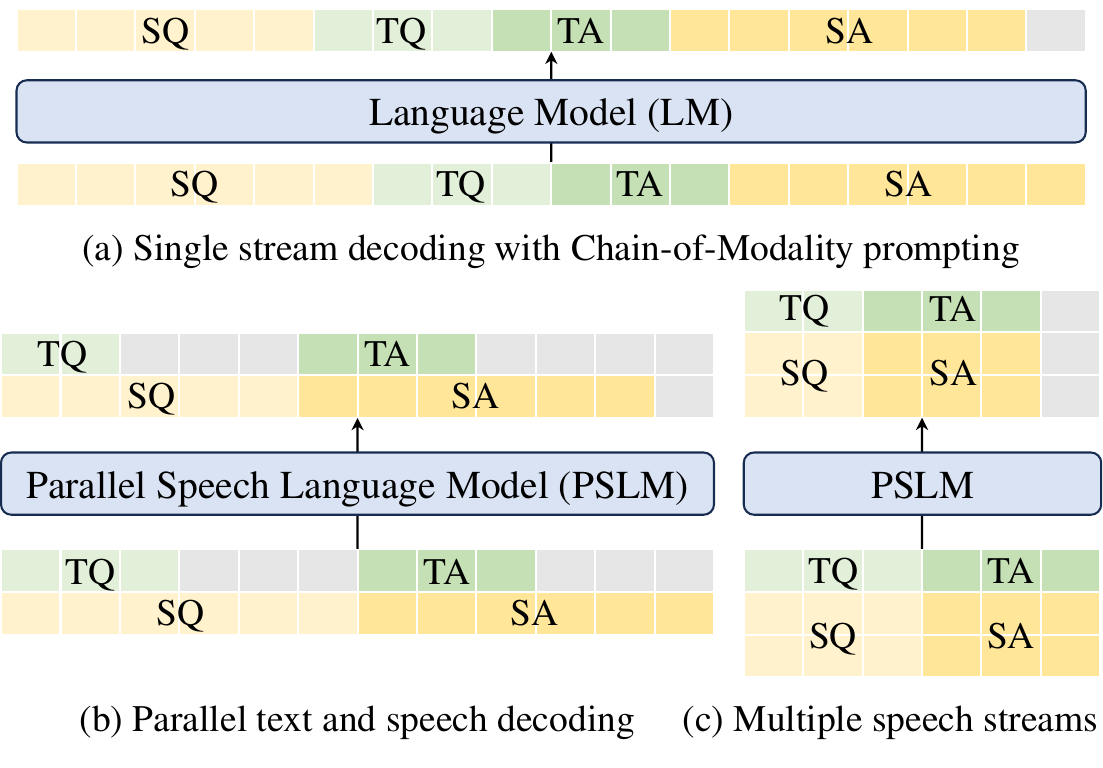}
  \vspace{-10pt}
  \caption{(a) Chain-of-Modality prompting necessitates generating text questions (TQ) and text answers (TA) from speech questions (SQ) before producing speech answers (SA). (b) Our Parallel Speech Language Model (\prop) enables the parallel decoding of TA and SA, reducing overall latency. (c) Introducing multiple speech streams further accelerates the generation of SA.}
  \label{fig:overview}
  \vspace{-10pt}
\end{figure}

\citet{zhang-etal-2023-speechgpt} proposed SpeechGPT, an LLM that receives speech questions (SQ) as speech tokens, which are discrete representations extracted from raw waveforms, and sequentially generates text questions (TQ), text answers (TA), and speech answers (SA).
\Figref{fig:overview} (a) illustrates their approach called Chain-of-Modality (CoM) prompting.
Spectron~\citep{nachmani2024spoken} follows this prompting style but directly handles speech spectrograms.
Although these methods can generate high-quality responses, they face two major challenges in terms of response latency.
First, generating SA requires the prior generation of TQ and TA.
Second, speech sequences are much longer than text sequences\footnote{Actual sequence lengths are provided in \Appref{app:seq_len}.}.

In this study, we propose Parallel Speech Language Model (PSLM), an LLM with multiple input-output sequences to handle both text and speech tokens, enabling their parallel generation.
To emphasize their parallel processing capabilities, we will refer to these sequences as ``streams''.
As described in \Figref{fig:overview} (b), PSLM begins to generate SA immediately after the end of SQ tokens, which can reduce overall latency.
This leads to our first research question (\textbf{RQ1}): Can PSLM improve latency while maintaining the response quality achieved by CoM prompting?
Additionally, we address the second challenge by introducing multiple speech streams to decode multiple speech tokens in a single step, as described in \Figref{fig:overview} (c).
This brings us to the second research question (\textbf{RQ2}): Do multiple speech streams sacrifice the response quality?
Addressing these questions will pave the way for more advanced and responsive applications of spoken dialogue systems.

\section{PSLM}

\subsection{Speech Discretization}
\label{sec:preliminary}

\paragraph{Speech Tokenization}
Extracting discrete speech tokens from raw waveforms enables language models to handle speech in the same manner as text tokens. 
Self-supervised learning has been widely used for speech tokenization due to its ability to extract spoken content from raw waveforms~(e.g., \citealt{rubenstein2023audiopalm,chou-etal-2023-toward,hassid2023twist}).
Following \citet{zhang-etal-2023-speechgpt}, we employ Hidden-Unit BERT (HuBERT)~\citep{hsu2021hubert} for speech tokenization.

\paragraph{Speech Detokenization}
In contrast to text tokenization, which is uniquely recoverable, speech tokenization largely discards the information of raw waveforms.
Two major approaches have been proposed to solve this problem.
The first approach uses a neural vocoder for directly reconstructing raw waveforms from speech tokens~(e.g., \citealt{zhang-etal-2023-speechgpt,chou-etal-2023-toward,hassid2023twist}).
The second approach uses a pretrained neural audio codec, which requires an additional module to predict the codec's tokens~(e.g., \citealt{rubenstein2023audiopalm,zhang2024speechgptgen}).
We adopt the first approach to reduce overall latency using HiFi-GAN~\citep{kong2020hifigan}, a non-autoregressive neural vocoder that efficiently generates high-fidelity waveforms.

\subsection{Integrating LMs with a Speech Stream}

\begin{figure}[t]
  \includegraphics[width=\columnwidth]{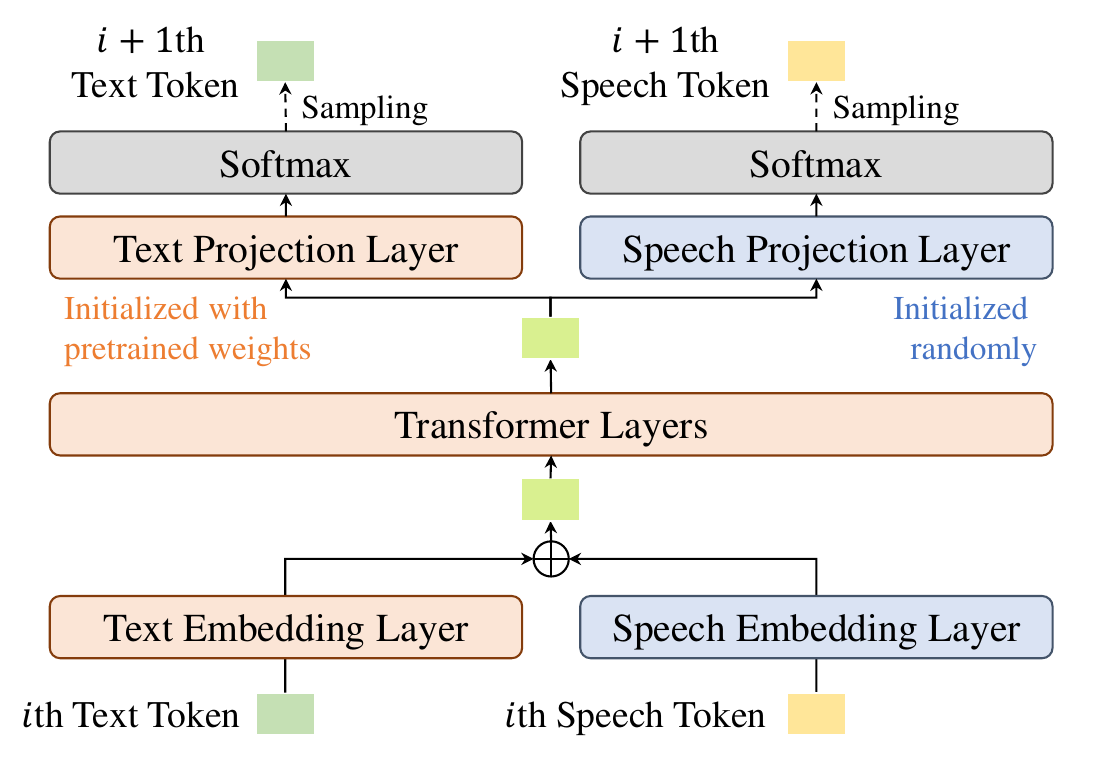}
  \vspace{-20pt}
  \caption{\revise{Architecture of PSLM.}}
  \label{fig:architecture}
  \vspace{-10pt}
\end{figure}

PSLM is built on top of a pretrained decoder-only Transformer~\citep{vaswani2017transformer}.
\revise{An overview of the PSLM architecture is provided in \Figref{fig:architecture}.}
We add new input embedding and output projection layers to process speech tokens, while the structure of the intermediate Transformer layers remains unchanged.
\revise{The embeddings of text and speech tokens are summed before being fed to the Transformer layers.
The hidden features from the final Transformer layer are passed to two output projection layers to calculate the logits of the next text and speech tokens.}
We randomly initialize the weights of new embedding and projection layers.

A challenge of joint text-speech modeling lies in the mismatch in their lengths.
In this study, we simply right-pad TQ and TA sequences with special \texttt{[TEXT-PAD]} tokens to align their lengths with those of the SQ and SA sequences, respectively.
\revise{
In a preliminary experiment on the CoM-based architecture, we attempted to generate text tokens and their corresponding speech tokens alternatively in a similar manner to ELLA-V~\citep{song2024ellav}; however, this approach led to frequent mispronunciation.
This is mainly because, in our case, the text is represented by tokens rather than phonemes; in some languages, the pronunciation of a character often changes according to subsequent characters, and a certain amount of lookahead is necessary to achieve accurate pronunciation.
In contrast, our alignment strategy allows the model to focus on text token generation initially and then refer to the generated text when producing the majority of speech tokens, leading to more accurate pronunciation.
} 

Our \prop is trained by minimizing the sum of cross entropy losses for each stream.
We include prompt tokens, comprising TQ and SQ, in the loss calculation.
During inference, \prop receives these prompt tokens and generates TA and SA in parallel.
\revise{Text and speech tokens are sampled independently from their respective distributions.}

\subsection{Introducing Multiple Speech Streams}

For further acceleration, we introduce multiple speech streams to \prop.
Assume that \prop has $1 + S$ streams, one for text tokens and $S$ for speech tokens.
\revise{Given the original speech token sequence of length $N$, the $s$-th speech stream consists of the speech tokens with indices $s, s+S, s+2S, ..., s+MS$, where $s\in \{1,\ldots,S\}$ and $M=\lfloor N/S \rfloor - 1$.
Compared to simply increasing the batch size, where the system's throughput improves but the latency for each instance remains unchanged, our approach reduces the sequence length handled by the Transformer layers to $1/S$, leading to an approximate $S$-fold speedup even in the single-instance scenario.}

During training, simply summing the cross entropy losses for each stream makes the loss of text tokens less dominant, leading to poor text generation quality.
Therefore, we introduce a weighted loss, where we multiply the loss for speech streams by $1/S$ to balance the weight of losses for text and speech streams.

\subsection{Streaming Inference with HiFi-GAN}
Following \citet{chen2022streaming}, we use HiFi-GAN for streaming inference; specifically, we provide partial speech tokens to generate waveform fragments.
In this study, we use non-causal convolution to maintain high speech quality.
Therefore, the first speech fragment can be generated once $N_\textrm{offset} = \lfloor R / 2 \rfloor + 1$ tokens are decoded, where $R$ denotes the receptive field of HiFi-GAN.
Implementation details can be found in \Appref{app:hifigan}.

\subsection{Overall Latency}
\label{sec:latency}
We define latency as the delay between the end of the user's utterance and the system's initial response.
The latency of conventional CoM-based systems $L_\textrm{CoM}$ can be represented as follows:
\begin{align}
    L_\textrm{CoM} &= D_\textrm{s2t} + D_\textrm{SQ} + \frac{N_\textrm{dec}}{P} + D_\textrm{t2s} \\
    N_\textrm{dec} &= N_\textrm{TQ} + N_\textrm{TA} + N_\textrm{offset}
    \label{eq:latency_com}
\end{align}
where $D_\textrm{s2t}$, $D_\textrm{SQ}$, and $D_\textrm{t2s}$ denote the delays of speech tokenization, the prefill phase in LMs, and speech detokenization, respectively; $N_\textrm{TQ}$ and $N_\textrm{TA}$ denote the number of tokens in TQ and TA, respectively; and $P$ denotes the tokens per second (TPS) during the decode phase in LMs.

Our \prop eliminates the need for generating TQ and TA beforehand, although it requires to run external ASR to obtain TQ.
Hence, its latency $L_\textrm{\prop}$ can be represented as follows:
\begin{align}
    L_\textrm{\prop} = D_\textrm{ASR} + D_\textrm{SQ} + \frac{N_\textrm{offset}}{P\cdot S} + D_\textrm{t2s}
    \label{eq:latency_pslm}
\end{align}
where $D_\textrm{ASR}$ denotes the ASR delay.
Here $D_\textrm{s2t}$ is omitted because speech tokenization can be performed in parallel with ASR.

\section{Experimental Setup}
\subsection{Dataset}
We used an internal dataset comprising 1.8M written QA pairs for training \revise{all models.}
\revise{Since some of these samples, which were primarily crawled from the internet, were deemed unsuitable for evaluation, we used a publicly available Japanese dataset~\citep{megagonlabs_instruction_ja} for evaluation.
This dataset was manually reviewed and consists of 669 diverse written QA pairs.
We further filtered the evaluation set by excluding samples whose TQ or TA exceeded 140 characters, the maximum number of characters observed in the training set.
The final evaluation set contained 396 samples.}
For both \revise{the training and evaluation} sets, we constructed a spoken question answering (SQA) dataset by synthesizing SQ and SA using a well-trained single-speaker TTS system based on VITS~\citep{kim2021vits}.

\subsection{Configuration}
\label{sec:config}
\paragraph{Tokenization and Detokenization}
For text tokenization, we used the tokenizer with a vocabulary size of 151,936 from rinna/nekomata-7b\footnote{\url{https://huggingface.co/rinna/nekomata-7b}}.
For speech tokenization, we applied $k$-means clustering with $k=512$ to 12-th layer features from rinna/japanese-hubert-base\footnote{\url{https://huggingface.co/rinna/japanese-hubert-base}}~\citep{sawada2024release}, obtaining 50 speech tokens per second.
For speech detokenization, we trained discrete unit-based HiFi-GAN~\citep{polyak2021speech} using pairs of synthesized speech waveforms of SQ and SA and their corresponding speech tokens.
For ASR, Whisper large-v3~\citep{radford2023robust} with faster-whisper\footnote{\url{https://github.com/SYSTRAN/faster-whisper}} was used throughout our experiments.

\paragraph{Language Modeling}
We used rinna/nekomata-7b, a 32-layer 4096-hidden-size Transformer LM that was continuously pretrained from Qwen-7B~\citep{bai2023qwen} on Japanese text, as the backbone of our models.
We implemented our models using the GPT-NeoX library~\citep{gpt-neox-library}.
Unless otherwise noted, models were trained for 50k steps with a batch size of 16 on 8 NVIDIA A100 GPUs using an Adam optimizer~\citep{kingma2015adam} with a peak learning rate set to 1e-5.
During inference, we set the temperature to 0.8 and applied top-$k$ and top-$p$ sampling with $k=60$ and $p=0.8$.

\subsection{Baselines}

\label{sec:baselines}
We involved three CoM-based baselines, which share the model weights but differ in their prompts during decoding: (1) \method{CoM-SQ} receives only SQ, (2) \method{CoM-ASR} receives SQ and transcribed TQ, and (3) \method{CoM} receives SQ and gold TQ.
\revise{In our preliminary experiments, the three-stage training~\citep{zhang-etal-2023-speechgpt} was not effective in our configuration; thus, we trained the model using the same configuration as described in \Secref{sec:config}.}

\subsection{Evaluation Metrics}
\label{sec:metrics}
\paragraph{ChatGPT Scores}
We used OpenAI's GPT-3.5 Turbo API to evaluate response quality on a 5-point scale from 1 (bad) to 5 (excellent).
The prompt is described in \Appref{app:prompt}.
We report the scores for TA and the transcription of SA as T-score and S-score, respectively.

\paragraph{Character Error Rate (CER)}
We calculated the character error rate between the generated TA and the transcription of SA to assess their alignment.

\paragraph{Failure Rate (FR)}
We counted failure cases such as (1) no \texttt{[EOS]} token was generated before the total sequence length reached 2048, or (2) tokens were generated in the wrong modality, i.e., speech tokens in TQ and TA, or text tokens in SA. 

\paragraph{Latency}
We simulated latency according to Equations \ref{eq:latency_com} and \ref{eq:latency_pslm} for each sample in the evaluation set, and reported the median values.
\revise{We set $D_\textrm{s2t} = 0.05$, $D_\textrm{SQ} = 0.05$, $D_\textrm{ASR} = 0.2$, and $D_\textrm{t2s} = 0.01$ based on measurements taken on a single NVIDIA A100 GPU.
For the TPS value $P$, the actual TPS varies depending on computing resources and optimization; 70 TPS was achieved with vLLM~\citep{kwon2023vllm} optimization, and 25 TPS without it.
Meanwhile, for streaming inference with HiFi-GAN, LMs need to generate 50 speech tokens per second.
Therefore, we set $P$ to 50 in our simulations to match this requirement.}

\input{tab/eval}

\paragraph{Human Rating}
We also conducted two subjective evaluations: one for text and the other for speech.
In the text evaluation, we presented pairs of gold TQ and generated TA, \revise{and raters evaluated the naturalness of TA based on the same criteria used in the ChatGPT-based evaluation (Text Naturalness).}
In the speech evaluation, we presented gold SQ and generated SA successively, along with their TQ and TA, and asked the raters to evaluate (1) how natural the SA is as the speech of the TA \revise{(Speech Naturalness)}, and (2) whether the response is fast enough \revise{(Speed Score)}.
\revise{For better reproducibility, we provide the actual instruction used for speech evaluation in \Appref{app:speecheval_instruction}.}
The duration of silence between SQ and SA was simulated in the manner described in \Secref{sec:latency}, except for the Ground Truth where the silence duration was set to 200ms, the average turn-taking gap in human conversation~\citep{levinson2015timing}.
Scores were rated on a 5-point scale.
Fifty samples were randomly chosen from the evaluation set, and twenty in-house workers rated twenty samples each.

\section{Results and Discussion}
\subsection{Automatic Evaluation}
\label{sec:auto_eval}

\paragraph{Comparison with Baselines}
To answer \textbf{RQ1}, we compared the proposed method in two conditions, \method{\prop} and \method{\prop-ASR}, with the baselines described in \Secref{sec:baselines}.
\method{\prop} receives SQ and gold TQ, while \method{\prop-ASR} receives SQ and transcribed TQ.
\Tblref{tbl:auto_eval} summarizes the results.
When gold TQ was given, \method{\prop} achieved comparable scores to \method{CoM} and significantly improved latency.
A similar trend was observed under more practical conditions where gold TQ was not available (\method{\prop-ASR} vs. \method{CoM-ASR}).
However, their scores were lower than those with gold TQ, and \method{CoM-SQ} faced greater degradation.
These results suggest that ASR performance is crucial for response quality, and \method{CoM-SQ} seems to have produced more ASR errors than Whisper.
Nevertheless, we conclude that PSLM maintains the response quality of CoM (\textbf{RQ1}).
We also found that \prop-based methods achieved lower FRs than CoM-based ones.
Each stream of \prop is dedicated to a single modality, which could have reduced the failures in generation.
Furthermore, methods other than \method{CoM-SQ} marked lower CERs than \method{Ground Truth}.
From this result, we confirmed that both CoM and \prop can generate appropriate SA corresponding to TA.

\paragraph{Multiple Speech Streams}
To answer \textbf{RQ2}, we trained \prop variants with two (\method{-2x}) or three (\method{-3x}) speech streams\footnote{\method{PSLM-3x} was trained with a batch size of 4 due to the increased number of parameters.}.
\method{PSLM-2x} achieved comparable scores to \method{PSLM}, whereas \method{PSLM-3x} demonstrated significant degradation.
From these results, we conclude that speech tokens can be decoded in up to two streams without quality degradation (\textbf{RQ2}).
An ablation study can be found in \Appref{app:ablation}.

\subsection{Human Evaluation}
Considering practical applicability to SQA, we manually evaluated three methods: \method{CoM-SQ}, \method{CoM-ASR}, and \method{\prop-ASR}, which do not rely on gold TQ, along with \method{Ground-Truth}.
\Tblref{tbl:human_eval} shows the results.
The text response naturalness of \method{\prop-ASR} was comparable to \method{CoM-ASR} and higher than \method{CoM-SQ}, which is consistent with the automatic evaluation results.
For speech naturalness, all methods achieved higher scores than \method{Ground-Truth}.
This result can be attributed to two reasons: (1) SA of \method{Ground-Truth} are synthetic speech, which may include errors in pronunciation, intonation, and pauses, and (2) SA of \method{Ground-Truth} are typically longer than those of other methods, incurring that one or two unnatural parts lowered the entire score.
Nevertheless, we confirmed that our approach can generate natural and faithful speech responses.
For response speed evaluation, \method{\prop-ASR} achieved a significantly higher score than \method{CoM-ASR} and \method{CoM-SQ}.
This finding verifies that the proposed method reduces latency both numerically and perceptibly.
Detailed analysis can be found in the next subsection.

\input{tab/human_eval}

\subsection{Detailed Latency Analysis}
\begin{figure}[t]
    \begin{center}
  \includegraphics[width=\columnwidth]{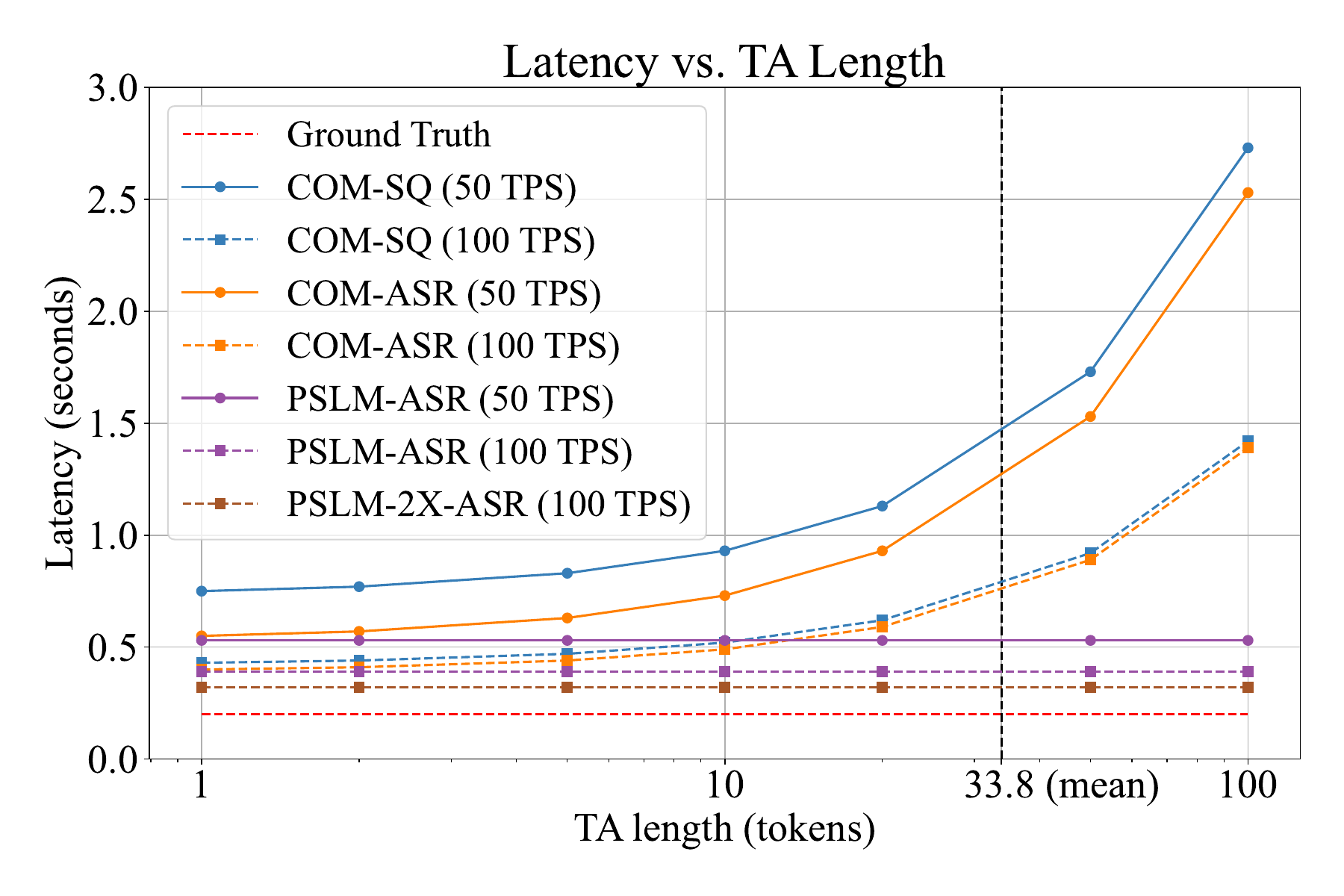}
    \vspace{-20pt}
    \caption{Latency vs. TA length for different methods and tokens per second (TPS). \prop-2x-ASR (50 TPS) is omitted because its latency is identical to \prop-ASR (100 TPS).}
    \label{fig:latency_vs_TAlen}
    \end{center}
    \vspace{-10pt}
\end{figure}

\revise{
The sequence length of TA, or $N_\textrm{TA}$, is the most influential factor in overall latency of CoM-based systems, as TA must be generated before SA. 
Thus, we investigated the overall latency by varying $N_\textrm{TA}$.}
\Figref{fig:latency_vs_TAlen} shows the results.
Due to the need for prior generation of TA, the latency of \method{CoM-SQ} and \method{CoM-ASR} increases linearly as TA length increases.
In contrast, the latency of \method{\prop-ASR} is constant because \Eqref{eq:latency_pslm} does not include $N_\textrm{TA}$, and \method{\prop-2x-ASR} further reduces the latency.
The gap between CoM-based and \prop-based systems is remarkable when generating long TA, highlighting the effectiveness of generating text and speech tokens in parallel.

\section{Conclusion}
In this study, we proposed the Parallel Speech Language Model (PSLM), an LLM capable of generating text and speech tokens in parallel with multiple input-output streams, and investigated its impact on response quality and overall latency.
The experimental evaluations on spoken question answering demonstrated that the proposed method significantly reduces latency compared to existing methods while maintaining response quality.
Future work includes verifying the effectiveness of the proposed method on larger datasets and real speech data.
Additionally, extending the proposed method to multi-turn dialogues is an important research direction.

\section{Limitations}
We recognize several limitations of this study.
First, \prop sacrifices ASR capability for faster response, requiring an external ASR module to serve as a spoken dialogue system.
\revise{Although this dependency can complicate the system structure, it does not degrade the system's performance, provided that an appropriate ASR module is selected.
This is supported by the fact that CoM-ASR outperformed CoM-SQ, as described in \Secref{sec:auto_eval}.} 
Nevertheless, enabling ASR with the PSLM architecture can be an interesting research direction.
Second, we used single-speaker synthetic speech for SQ and SA, which lacks diversity in several aspects of speech such as accent, rhythm, emotion, and timbre.
Practical applications may require to accept voices of arbitrary speakers, which we will address in future work.
Finally, multi-turn dialogue settings were not investigated in our experiments.
While SpeechGPT~\citep{zhang-etal-2023-speechgpt} was not applied to multi-turn dialogue due to sequence length limitations, our models with multiple speech streams have the potential to perform multi-turn dialogue.

\bibliography{anthology,ref/speech,ref/ml,ref/nlp}
\appendix
\input{appendix}

\end{document}

%% file: math_commands.tex
\usepackage{amsmath,amsfonts,bm}

\def\Tblref#1{Table~\ref{#1}}

\def\Figref#1{Figure~\ref{#1}}

\def\Secref#1{Section~\ref{#1}}

\def\Appref#1{Appendix~\ref{#1}}
\def\eqref#1{equation~\ref{#1}}
\def\Eqref#1{Equation~\ref{#1}}

\def\1{\bm{1}}

\DeclareMathAlphabet{\mathsfit}{\encodingdefault}{\sfdefault}{m}{sl}
\SetMathAlphabet{\mathsfit}{bold}{\encodingdefault}{\sfdefault}{bx}{n}

\newcommand{\method}[1]{#1}
\newcommand{\prop}{PSLM\xspace}

%% file: tab/eval.tex
\begin{table*}[t]
\caption{Automatic evaluation results. T-score and S-score represent the ChatGPT-based score for TA and transcribed SA, respectively. FR denotes the failure rate. Latency values in parentheses represent inputs involving gold TQ.}
\vspace{-5pt}
\label{tbl:auto_eval}
\begin{center}
\small
\begin{tabular}{l|l|l|ccccc}
\toprule
\textbf{Method} & \textbf{Input modality} & \textbf{Output Modality} & \textbf{T-score}$\uparrow$ & \textbf{S-score}$\uparrow$ & \textbf{FR}$\downarrow$ & \textbf{CER}$\downarrow$ & \textbf{Latency [s]}$\downarrow$\\
\midrule
\method{Ground Truth} & --- & --- & 4.00$\pm$0.02 & 3.58$\pm$0.06 & --- & 7.35 & ---\\ \midrule

\method{CoM} & SQ $\rightarrow$ TQ (Gold) & TA $\rightarrow$ SA & 3.50$\pm$0.09 & 3.27$\pm$0.09 & 12.12 & 6.28 & (0.67) \\
\method{\prop} & SQ, TQ (Gold) & TA, SA & 3.50$\pm$0.08 & 3.22$\pm$0.09 & 5.05 & 5.25 & (0.34) \\\midrule

\method{CoM-SQ} & SQ & TQ $\rightarrow$ TA $\rightarrow$ SA & 3.12$\pm$0.11 & 2.94$\pm$0.10 & 15.91 & 7.83 & 1.03 \\
\method{CoM-ASR} & SQ $\rightarrow$ TQ (ASR) & TA $\rightarrow$ SA & 3.27$\pm$0.10 & 3.07$\pm$0.09 & 13.13 & 6.18 & 0.92 \\
\method{\prop-ASR} & SQ, TQ (ASR) & TA, SA & 3.34$\pm$0.09 & 3.05$\pm$0.10 & 6.31 & 6.05 & 0.54 \\ \midrule

\method{\prop-2x} & SQ, TQ (Gold) & TA, SA & 3.50$\pm$0.08 & 3.20$\pm$0.09 & 4.29 & 6.39 & (0.20) \\
\method{\prop-3x} & SQ, TQ (Gold) & TA, SA & 3.28$\pm$0.10 & 2.99$\pm$0.10 & 7.07 & 6.09 & (0.15) \\
\bottomrule
\end{tabular}
\end{center}
\vspace{-5pt}
\end{table*}

%% file: tab/human_eval.tex
\begin{table}[t]
\caption{Human evaluation results.}
\vspace{-5pt}
\label{tbl:human_eval}
\begin{center}
\small
\begin{tabular}{l|ccc}
\toprule
\textbf{Method} & \textbf{Text}$\uparrow$ & \textbf{Speech}$\uparrow$ & \textbf{Speed}$\uparrow$\\
\midrule
\method{Ground Truth} & 4.08$\pm$0.26 & 3.74$\pm$0.19 & 4.73$\pm$0.11 \\
\method{CoM-SQ} & 2.44$\pm$0.29 & 4.04$\pm$0.20 & 4.07$\pm$0.23\\
\method{CoM-ASR} & 2.90$\pm$0.30 & 3.94$\pm$0.20 & 4.17$\pm$0.22\\
\method{\prop-ASR} & 3.08$\pm$0.27 & 4.08$\pm$0.20 & 4.57$\pm$0.13\\
\bottomrule
\end{tabular}
\end{center}
\vspace{-10pt}
\end{table}

%% file: appendix.tex
\newpage
\section{Sequence Length Distributions}
\label{app:seq_len}

We calculated the sequence length distributions of SQ, TQ, TA, and SA in the training set.
The results are listed in \Tblref{tbl:seq_len}.
On average, CoM prompting requires to generate $36.5 \ (\textrm{TQ}) + 33.8 \ (\textrm{TA}) \approx 70$ text tokens before generating SA.
Eliminating the need for generating these tokens can greatly reduce overall latency.
In addition, speech tokens are more than 11 times longer than text tokens, highlighting the need for efficient generation of speech tokens.

\section{Implementation Details of HiFi-GAN}
\label{app:hifigan}
\begin{figure}[t]
  \includegraphics[width=\columnwidth]{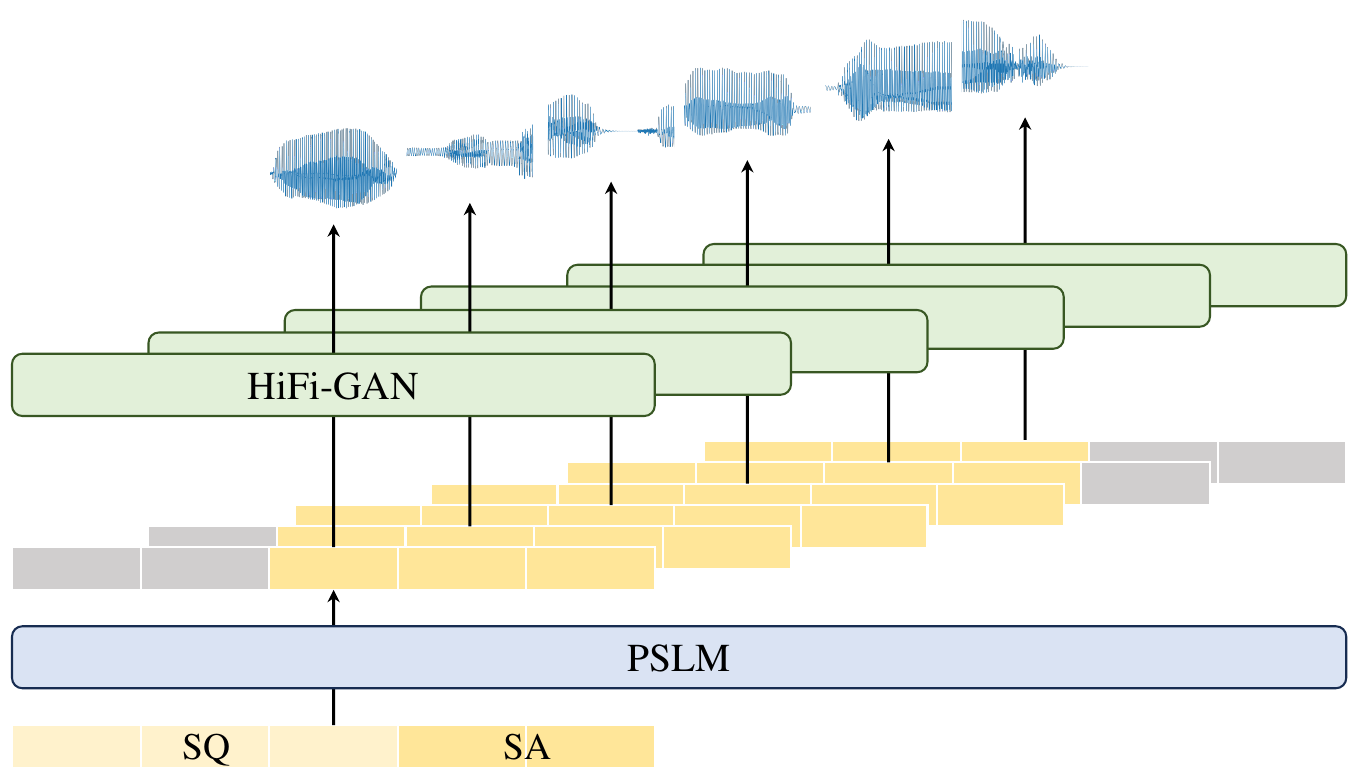}
  \vspace{-5pt}
  \caption{Streaming inference using HiFi-GAN with receptive field size $R=5$ and SA length $N_\textrm{SA}=6$. Waveform generation begins once $N_\textrm{offset} = \lfloor R / 2 \rfloor + 1 = 3$ tokens are generated. Text tokens are omitted.}
  \label{fig:hifigan}
  \vspace{-5pt}
\end{figure}

The HiFi-GAN generator comprises convolution layers.
Therefore, a waveform fragment corresponding to the $i$-th token depends only on tokens with indices $[i - \lfloor R / 2 \rfloor, i + \lfloor R / 2 \rfloor]$.
This allows waveform generation to start before the entire SA is generated.
As described in \Figref{fig:hifigan}, HiFi-GAN first generates a waveform fragment once the LM generates $N_\textrm{offset} = \lfloor R / 2 \rfloor + 1$ tokens, then generates subsequent fragments by shifting input tokens one by one.

In our experiments, we trained HiFi-GAN to generate 24~kHz waveform from 50Hz tokens, which results in $R=26$.
Following \citet{polyak2021speech}, we embedded input speech tokens into 256-dimensional features and fed them to HiFi-GAN.
We modified the upsampling rates to $[8, 6, 5, 2]$, the number of total iterations to 300k, and kept the other configuration the same as the original work~\citep{kong2020hifigan}.

\section{ChatGPT Evaluation Prompt}
\label{app:prompt}
\begin{figure*}[h]
  \begin{center}
  \includegraphics[width=0.8\linewidth]{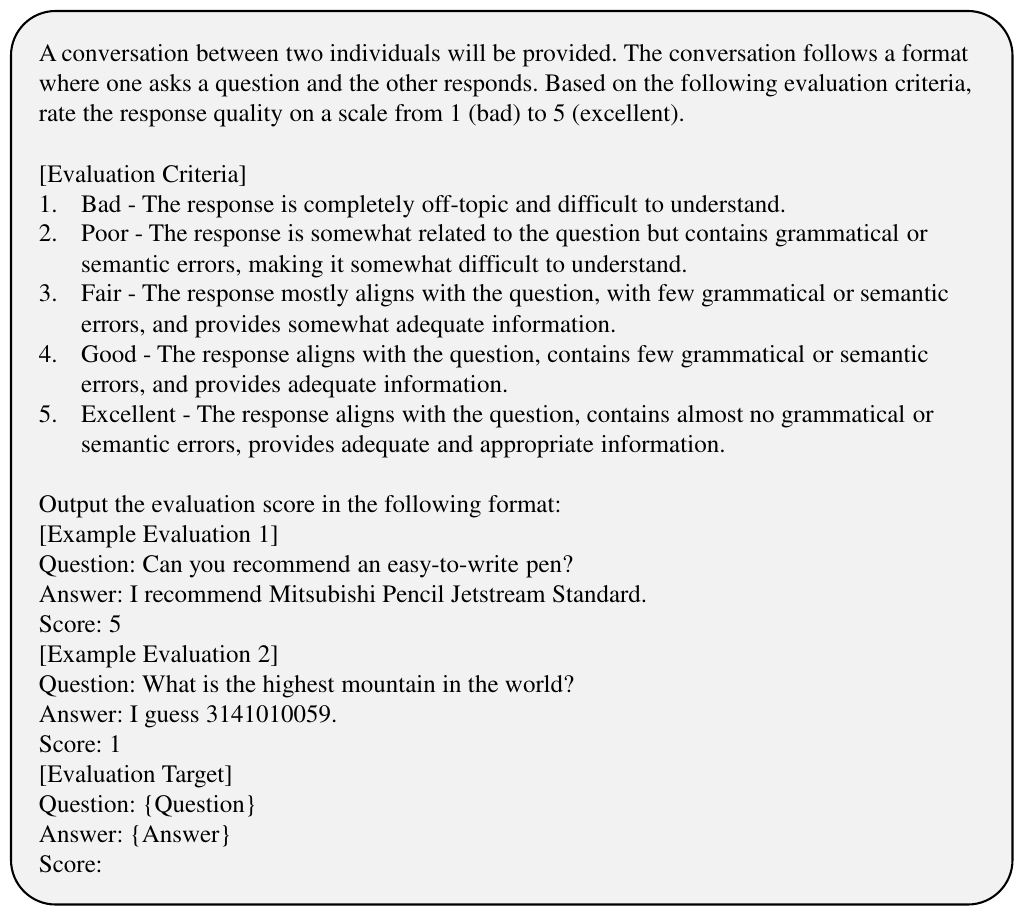}
  \vspace{-5pt}
  \caption{Prompt for ChatGPT evaluation.}
  \label{fig:prompt}
  \end{center}
  \vspace{-10pt}
\end{figure*}

We used the prompt in \Figref{fig:prompt} for ChatGPT-based evaluation.
The original prompt was written in Japanese, but a translated version is presented here.

\section{Speech Evaluation Instruction}
\label{app:speecheval_instruction}
\begin{figure*}[h]
  \begin{center}
  \includegraphics[width=0.8\linewidth]{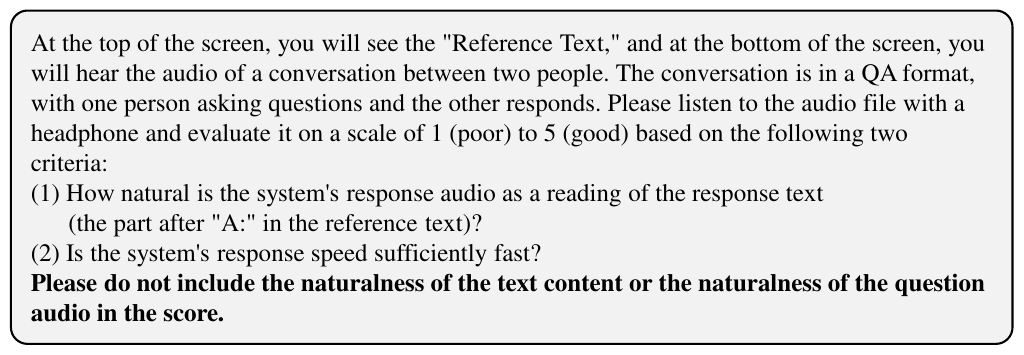}
  \vspace{-5pt}
  \caption{\revise{Instruction for speech evaluation.}}
  \label{fig:speecheval_instruction}
  \end{center}
  \vspace{-10pt}
\end{figure*}

\revise{
We used the instruction in \Figref{fig:speecheval_instruction} for speech evaluation.
The original instruction was written in Japanese, but a translated version is presented here.
}

\section{Ablation Study}
\label{app:ablation}

We trained three \prop variants, one from scratch (\method{-no-pretrain}), one without TQ (\method{-no-TQ}), and one without SQ (\method{-no-SQ}).
In addition, we trained \prop-2x and \prop-3x without weighted loss (\method{-no-WL}).
\Tblref{tbl:ablation} shows the automatic evaluation results.
\method{\prop-no-pretrain} exhibited significant degradation in all metrics, indicating the necessity of pretrained LM's text capability.
\method{\prop-no-TQ} also showed large degradation, highlighting the importance of TQ in response quality.
In contrast, \method{\prop-no-SQ} achieved comparable scores to \method{\prop}.
This result implies that the speech-specific information such as intonation, rhythm, and emotion is not essential in the current SQA task due to the use of synthetic speech.
We also found that \method{\prop-2x-no-WL} achieved almost comparable scores to \method{\prop}, whereas \method{\prop-3x-no-WL} showed significant degradation.
From these results, we conclude that the weighted loss is especially effective as the number of speech streams increases.

\input{tab/seq_len}

\input{tab/eval_ablation}

%% file: tab/seq_len.tex
\begin{table}[t]
\caption{Sequence length distributions in the training set (in tokens).}
\vspace{-10pt}
\label{tbl:seq_len}
\begin{center}
\small
\begin{tabular}{l|cccc}
\toprule
& \textbf{SQ} & \textbf{TQ} & \textbf{TA} & \textbf{SA}\\\midrule
Mean & 406.6 & 36.5 & 33.8 & 386.5\\\midrule
Min & 34 & 2 & 1 & 27\\
25\% & 214 & 19 & 15 & 179\\
50\% & 354 & 32 & 29 & 340\\
75\% & 577 & 51 & 50 & 563\\
Max & 1861 & 148 & 147 & 1697\\
\bottomrule
\end{tabular}
\end{center}
\vspace{0pt}
\end{table}

%% file: tab/eval_ablation.tex
\begin{table*}[b]
\caption{Ablation study. The suffix no-WL denotes weighted loss was not applied.}
\vspace{-10pt}
\label{tbl:ablation}
\begin{center}
\small
\begin{tabular}{l|l|l|cccc}
\toprule
\textbf{Method} & \textbf{Input modality} & \textbf{Output Modality} & \textbf{T-score}$\uparrow$ & \textbf{S-score}$\uparrow$ & \textbf{FR}$\downarrow$ & \textbf{CER}$\downarrow$\\
\midrule
\method{\prop} & SQ, TQ (Gold) & TA, SA & 3.50$\pm$0.08 & 3.22$\pm$0.09 & 5.05 & 5.25\\
\method{\prop-2x} & SQ, TQ (Gold) & TA, SA & 3.50$\pm$0.08 & 3.20$\pm$0.09 & 4.29 & 6.39\\
\method{\prop-3x} & SQ, TQ (Gold) & TA, SA & 3.28$\pm$0.10 & 2.99$\pm$0.10 & 7.07 & 6.09\\
\midrule
\method{\prop-no-pretrain} & SQ, TQ (Gold) & TA, SA & 2.22$\pm$0.07 & 2.12$\pm$0.07 & 18.18 & 10.13\\
\method{\prop-no-TQ} & SQ & TA, SA & 2.34$\pm$0.09 & 2.19$\pm$0.09 & 8.84 & 6.38 \\
\method{\prop-no-SQ} & TQ (Gold) & TA, SA & 3.54$\pm$0.08 & 3.17$\pm$0.09 & 6.31 & 8.99 \\
\method{\prop-2x-no-WL} & SQ, TQ (Gold) & TA, SA & 3.42$\pm$0.08 & 3.17$\pm$0.08 & 8.84 & 4.99 \\
\method{\prop-3x-no-WL} & SQ, TQ (Gold) & TA, SA & 2.67$\pm$0.10 & 2.46$\pm$0.10 & 11.36 & 6.94 \\
\bottomrule
\end{tabular}
\end{center}
\vspace{-10pt}
\end{table*}

%% file: arxiv.bbl
\begin{thebibliography}{30}
\providecommand{\natexlab}[1]{#1}

\bibitem[{Andonian et~al.(2023)Andonian, Anthony, Biderman, Black, Gali, Gao, Hallahan, Levy-Kramer, Leahy, Nestler, Parker, Pieler, Phang, Purohit, Schoelkopf, Stander, Songz, Tigges, Thérien, Wang, and Weinbach}]{gpt-neox-library}
Alex Andonian, Quentin Anthony, Stella Biderman, Sid Black, Preetham Gali, Leo Gao, Eric Hallahan, Josh Levy-Kramer, Connor Leahy, Lucas Nestler, Kip Parker, Michael Pieler, Jason Phang, Shivanshu Purohit, Hailey Schoelkopf, Dashiell Stander, Tri Songz, Curt Tigges, Benjamin Thérien, Phil Wang, and Samuel Weinbach. 2023.
\newblock \href {https://www.github.com/eleutherai/gpt-neox} {{GPT-NeoX: Large Scale Autoregressive Language Modeling in PyTorch}}.

\bibitem[{Bai et~al.(2023)Bai, Bai, Chu, Cui, Dang, Deng, Fan, Ge, Han, Huang, Hui, Ji, Li, Lin, Lin, Liu, Liu, Lu, Lu, Ma, Men, Ren, Ren, Tan, Tan, Tu, Wang, Wang, Wang, Wu, Xu, Xu, Yang, Yang, Yang, Yang, Yao, Yu, Yuan, Yuan, Zhang, Zhang, Zhang, Zhang, Zhou, Zhou, Zhou, and Zhu}]{bai2023qwen}
Jinze Bai, Shuai Bai, Yunfei Chu, Zeyu Cui, Kai Dang, Xiaodong Deng, Yang Fan, Wenbin Ge, Yu~Han, Fei Huang, Binyuan Hui, Luo Ji, Mei Li, Junyang Lin, Runji Lin, Dayiheng Liu, Gao Liu, Chengqiang Lu, Keming Lu, Jianxin Ma, Rui Men, Xingzhang Ren, Xuancheng Ren, Chuanqi Tan, Sinan Tan, Jianhong Tu, Peng Wang, Shijie Wang, Wei Wang, Shengguang Wu, Benfeng Xu, Jin Xu, An~Yang, Hao Yang, Jian Yang, Shusheng Yang, Yang Yao, Bowen Yu, Hongyi Yuan, Zheng Yuan, Jianwei Zhang, Xingxuan Zhang, Yichang Zhang, Zhenru Zhang, Chang Zhou, Jingren Zhou, Xiaohuan Zhou, and Tianhang Zhu. 2023.
\newblock \href {https://arxiv.org/abs/2309.16609} {Qwen technical report}.
\newblock \emph{Computing Research Repository}, arxiv:2309.16609.

\bibitem[{Chen et~al.(2017)Chen, Liu, Yin, and Tang}]{hongshen2017survey}
Hongshen Chen, Xiaorui Liu, Dawei Yin, and Jiliang Tang. 2017.
\newblock \href {https://doi.org/10.1145/3166054.3166058} {A survey on dialogue systems: Recent advances and new frontiers}.
\newblock \emph{SIGKDD Explorations Newsletter}, 19(2):25--35.

\bibitem[{Chen et~al.(2022)Chen, Miao, and Zhang}]{chen2022streaming}
Ziyi Chen, Haoran Miao, and Pengyuan Zhang. 2022.
\newblock \href {https://arxiv.org/abs/2206.07288} {Streaming non-autoregressive model for any-to-many voice conversion}.
\newblock \emph{Computing Research Repository}, arXiv:2206.07288.

\bibitem[{Chou et~al.(2023)Chou, Chien, Hsu, Livescu, Babu, Conneau, Baevski, and Auli}]{chou-etal-2023-toward}
Ju-Chieh Chou, Chung-Ming Chien, Wei-Ning Hsu, Karen Livescu, Arun Babu, Alexis Conneau, Alexei Baevski, and Michael Auli. 2023.
\newblock \href {https://doi.org/10.18653/v1/2023.findings-emnlp.438} {Toward joint language modeling for speech units and text}.
\newblock In \emph{Findings of the Association for Computational Linguistics: EMNLP 2023}, pages 6582--6593, Singapore. Association for Computational Linguistics.

\bibitem[{Fathullah et~al.(2024)Fathullah, Wu, Lakomkin, Jia, Shangguan, Li, Guo, Xiong, Mahadeokar, Kalinli, Fuegen, and Seltzer}]{fathullah2024prompting}
Yassir Fathullah, Chunyang Wu, Egor Lakomkin, Junteng Jia, Yuan Shangguan, Ke~Li, Jinxi Guo, Wenhan Xiong, Jay Mahadeokar, Ozlem Kalinli, Christian Fuegen, and Mike Seltzer. 2024.
\newblock \href {https://doi.org/10.1109/ICASSP48485.2024.10447605} {Prompting large language models with speech recognition abilities}.
\newblock In \emph{Proceedings of the 2024 IEEE International Conference on Acoustics, Speech, and Signal Processing}, pages 13351--13355, Seoul, Korea.

\bibitem[{Hao et~al.(2023)Hao, Zhou, Liu, Li, Hu, Wang, and Wei}]{hao2023boosting}
Hongkun Hao, Long Zhou, Shujie Liu, Jinyu Li, Shujie Hu, Rui Wang, and Furu Wei. 2023.
\newblock \href {https://arxiv.org/abs/2401.00246} {Boosting large language model for speech synthesis: An empirical study}.
\newblock \emph{Computing Research Repository}, arXiv:2401.00246.

\bibitem[{Hassid et~al.(2023)Hassid, Remez, Nguyen, Gat, Conneau, Kreuk, Copet, Defossez, Synnaeve, Dupoux, Schwartz, and Adi}]{hassid2023twist}
Michael Hassid, Tal Remez, Tu~Anh Nguyen, Itai Gat, Alexis Conneau, Felix Kreuk, Jade Copet, Alexandre Defossez, Gabriel Synnaeve, Emmanuel Dupoux, Roy Schwartz, and Yossi Adi. 2023.
\newblock \href {https://papers.nips.cc/paper_files/paper/2023/hash/c859b99b5d717c9035e79d43dfd69435-Abstract-Conference.html} {Textually pretrained speech language models}.
\newblock In \emph{Proceedings of the 37th International Conference on Neural Information Processing Systems}, pages 63483--63501, New Orleans, LA, U.S.A.

\bibitem[{Hayashibe(2023)}]{megagonlabs_instruction_ja}
Yuta Hayashibe. 2023.
\newblock \href {https://github.com/megagonlabs/instruction_ja} {{megagonlabs/instruction\_ja}: Japanese instructions data for {LLM}}.

\bibitem[{Hono et~al.(2024)Hono, Mitsuda, Zhao, Mitsui, Wakatsuki, and Sawada}]{hono-etal-2024-integrating}
Yukiya Hono, Koh Mitsuda, Tianyu Zhao, Kentaro Mitsui, Toshiaki Wakatsuki, and Kei Sawada. 2024.
\newblock \href {https://doi.org/10.18653/v1/2024.findings-acl.787} {Integrating pre-trained speech and language models for end-to-end speech recognition}.
\newblock In \emph{Findings of the Association for Computational Linguistics ACL 2024}, pages 13289--13305, Bangkok, Thailand and virtual meeting. Association for Computational Linguistics.

\bibitem[{Hsu et~al.(2021)Hsu, Bolte, Tsai, Lakhotia, Salakhutdinov, and Mohamed}]{hsu2021hubert}
Wei-Ning Hsu, Benjamin Bolte, Yao-Hung~Hubert Tsai, Kushal Lakhotia, Ruslan Salakhutdinov, and Abdelrahman Mohamed. 2021.
\newblock \href {https://doi.org/10.1109/TASLP.2021.3122291} {{HuBERT}: Self-supervised speech representation learning by masked prediction of hidden units}.
\newblock \emph{IEEE/ACM Transactions on Audio, Speech, and Language Processing}, 29:3451--3460.

\bibitem[{Jokinen and McTear(2009)}]{jokinen2009spoken}
Kristiina Jokinen and Michael McTear. 2009.
\newblock \href {https://doi.org/10.2200/S00204ED1V01Y200910HLT005} {\emph{Spoken Dialogue Systems}}.
\newblock Synthesis Lectures on Human Language Technologies. Morgan \& Claypool publishers, U.S.A.

\bibitem[{Kim et~al.(2021)Kim, Kong, and Son}]{kim2021vits}
Jaehyeon Kim, Jungil Kong, and Juhee Son. 2021.
\newblock \href {https://proceedings.mlr.press/v139/kim21f.html} {Conditional variational autoencoder with adversarial learning for end-to-end text-to-speech}.
\newblock In \emph{Proceedings of the 38th International Conference on Machine Learning}, pages 5530--5540, online.

\bibitem[{Kingma and Ba(2015)}]{kingma2015adam}
Diederik~P Kingma and Jimmy Ba. 2015.
\newblock \href {https://arxiv.org/abs/1412.6980} {Adam: A method for stochastic optimization}.
\newblock In \emph{Proceedings of the 3rd International Conference on Learning Representations}, San Diego, CA, U.S.A.

\bibitem[{Kong et~al.(2020)Kong, Kim, and Bae}]{kong2020hifigan}
Jungil Kong, Jaehyeon Kim, and Jaekyoung Bae. 2020.
\newblock \href {https://papers.nips.cc/paper_files/paper/2020/hash/c5d736809766d46260d816d8dbc9eb44-Abstract.html} {{HiFi-GAN}: Generative adversarial networks for efficient and high fidelity speech synthesis}.
\newblock In \emph{Proceedings of the 34th International Conference on Neural Information Processing Systems}, pages 17022--17033, online.

\bibitem[{Kwon et~al.(2023)Kwon, Li, Zhuang, Sheng, Zheng, Yu, Gonzalez, Zhang, and Stoica}]{kwon2023vllm}
Woosuk Kwon, Zhuohan Li, Siyuan Zhuang, Ying Sheng, Lianmin Zheng, Cody~Hao Yu, Joseph Gonzalez, Hao Zhang, and Ion Stoica. 2023.
\newblock \href {https://doi.org/10.1145/3600006.3613165} {Efficient memory management for large language model serving with pagedattention}.
\newblock In \emph{Proceedings of the 29th Symposium on Operating Systems Principles}, pages 611--626, New York, NY, U.S.A.

\bibitem[{Levinson and Torreira(2015)}]{levinson2015timing}
Stephen~C Levinson and Francisco Torreira. 2015.
\newblock \href {https://doi.org/10.3389/fpsyg.2015.00731} {Timing in turn-taking and its implications for processing models of language}.
\newblock \emph{Frontiers in psychology}, 6.

\bibitem[{McTear(2002)}]{mctear2002spoken}
Michael~F. McTear. 2002.
\newblock \href {https://doi.org/10.1145/505282.505285} {Spoken dialogue technology: enabling the conversational user interface}.
\newblock \emph{ACM Computing Surveys}, 34(1):90--169.

\bibitem[{Nachmani et~al.(2024)Nachmani, Levkovitch, Hirsch, Salazar, Asawaroengchai, Mariooryad, Rivlin, Skerry-Ryan, and Ramanovich}]{nachmani2024spoken}
Eliya Nachmani, Alon Levkovitch, Roy Hirsch, Julian Salazar, Chulayuth Asawaroengchai, Soroosh Mariooryad, Ehud Rivlin, RJ~Skerry-Ryan, and Michelle~Tadmor Ramanovich. 2024.
\newblock \href {https://arxiv.org/abs/2305.15255} {Spoken question answering and speech continuation using spectrogram-powered {LLM}}.
\newblock In \emph{In Proceedings of the 12th International Conference on Learning Representations}, Vienna, Austria.

\bibitem[{Polyak et~al.(2021)Polyak, Adi, Copet, Kharitonov, Lakhotia, Hsu, Mohamed, and Dupoux}]{polyak2021speech}
Adam Polyak, Yossi Adi, Jade Copet, Eugene Kharitonov, Kushal Lakhotia, Wei-Ning Hsu, Abdelrahman Mohamed, and Emmanuel Dupoux. 2021.
\newblock \href {https://doi.org/10.21437/Interspeech.2021-475} {Speech resynthesis from discrete disentangled self-supervised representations}.
\newblock In \emph{Proceedings of INTERSPEECH 2021}, pages 3615--3619, online.

\bibitem[{Radford et~al.(2023)Radford, Kim, Xu, Brockman, McLeavey, and Sutskever}]{radford2023robust}
Alec Radford, Jong~Wook Kim, Tao Xu, Greg Brockman, Christine McLeavey, and Ilya Sutskever. 2023.
\newblock \href {https://dl.acm.org/doi/10.5555/3618408.3619590} {Robust speech recognition via large-scale weak supervision}.
\newblock In \emph{Proceedings of the 40th International Conference on Machine Learning}, pages 28492--28518, Honolulu, Hawaii, U.S.A.

\bibitem[{Rubenstein et~al.(2023)Rubenstein, Asawaroengchai, Nguyen, Bapna, Borsos, Quitry, Chen, El~Badawy, Han, Kharitonov, Muckenhirn, Padfield, Qin, Rozenberg, Sainath, Schalkwyk, Sharifi, Ramanovich, Tagliasacchi, Tudor, Velimirović, Vincent, Yu, Wang, Zayats, Zeghidour, Zhang, Zhang, Zilka, and Frank}]{rubenstein2023audiopalm}
Paul~K. Rubenstein, Chulayuth Asawaroengchai, Duc~Dung Nguyen, Ankur Bapna, Zalán Borsos, Félix de~Chaumont Quitry, Peter Chen, Dalia El~Badawy, Wei Han, Eugene Kharitonov, Hannah Muckenhirn, Dirk Padfield, James Qin, Danny Rozenberg, Tara Sainath, Johan Schalkwyk, Matt Sharifi, Michelle~Tadmor Ramanovich, Marco Tagliasacchi, Alexandru Tudor, Mihajlo Velimirović, Damien Vincent, Jiahui Yu, Yongqiang Wang, Vicky Zayats, Neil Zeghidour, Yu~Zhang, Zhishuai Zhang, Lukas Zilka, and Christian Frank. 2023.
\newblock \href {https://arxiv.org/abs/2306.12925} {{AudioPaLM}: A large language model that can speak and listen}.
\newblock \emph{Computing Research Repository}, arXiv:2306.12925.

\bibitem[{Sawada et~al.(2024)Sawada, Zhao, Shing, Mitsui, Kaga, Hono, Wakatsuki, and Mitsuda}]{sawada2024release}
Kei Sawada, Tianyu Zhao, Makoto Shing, Kentaro Mitsui, Akio Kaga, Yukiya Hono, Toshiaki Wakatsuki, and Koh Mitsuda. 2024.
\newblock \href {https://aclanthology.org/2024.lrec-main.1213} {Release of pre-trained models for the {J}apanese language}.
\newblock In \emph{Proceedings of the 2024 Joint International Conference on Computational Linguistics, Language Resources and Evaluation}, pages 13898--13905, Torino, Italia. ELRA and ICCL.

\bibitem[{Song et~al.(2024)Song, Chen, Wang, Ma, and Chen}]{song2024ellav}
Yakun Song, Zhuo Chen, Xiaofei Wang, Ziyang Ma, and Xie Chen. 2024.
\newblock \href {https://arxiv.org/abs/2401.07333} {{ELLA-V}: Stable neural codec language modeling with alignment-guided sequence reordering}.
\newblock \emph{Computing Research Repository}, arXiv:2401.07333.

\bibitem[{Vaswani et~al.(2017)Vaswani, Shazeer, Parmar, Uszkoreit, Jones, Gomez, Kaiser, and Polosukhin}]{vaswani2017transformer}
Ashish Vaswani, Noam Shazeer, Niki Parmar, Jakob Uszkoreit, Llion Jones, Aidan~N Gomez, {\L}ukasz Kaiser, and Illia Polosukhin. 2017.
\newblock \href {https://papers.nips.cc/paper_files/paper/2017/hash/3f5ee243547dee91fbd053c1c4a845aa-Abstract.html} {Attention is all you need}.
\newblock In \emph{Proceedings of the 31st International Conference on Neural Information Processing Systems}, pages 5998--6008, Long Beach, CA, U.S.A.

\bibitem[{Wang et~al.(2023{\natexlab{a}})Wang, Wang, Du, Chen, Zhou, Wang, and Wong}]{wang2023survey}
Hongru Wang, Lingzhi Wang, Yiming Du, Liang Chen, Jingyan Zhou, Yufei Wang, and Kam-Fai Wong. 2023{\natexlab{a}}.
\newblock \href {https://arxiv.org/abs/2311.16789} {A survey of the evolution of language model-based dialogue systems}.
\newblock \emph{Computing Research Repository}, arxiv:2311.16789.

\bibitem[{Wang et~al.(2023{\natexlab{b}})Wang, Zhou, Zhang, Wu, Liu, Gaur, Chen, Li, and Wei}]{wang2023viola}
Tianrui Wang, Long Zhou, Ziqiang Zhang, Yu~Wu, Shujie Liu, Yashesh Gaur, Zhuo Chen, Jinyu Li, and Furu Wei. 2023{\natexlab{b}}.
\newblock \href {https://arxiv.org/abs/2305.16107} {{VioLA}: Unified codec language models for speech recognition, synthesis, and translation}.
\newblock \emph{Computing Research Repository}, arXiv:2305.16107.

\bibitem[{Yi et~al.(2024)Yi, Ouyang, Liu, Liao, Xu, and Shen}]{yi2024survey}
Zihao Yi, Jiarui Ouyang, Yuwen Liu, Tianhao Liao, Zhe Xu, and Ying Shen. 2024.
\newblock \href {https://arxiv.org/abs/2402.18013} {A survey on recent advances in llm-based multi-turn dialogue systems}.
\newblock \emph{Computing Research Repository}, arxiv:2402.18013.

\bibitem[{Zhang et~al.(2023)Zhang, Li, Zhang, Zhan, Wang, Zhou, and Qiu}]{zhang-etal-2023-speechgpt}
Dong Zhang, Shimin Li, Xin Zhang, Jun Zhan, Pengyu Wang, Yaqian Zhou, and Xipeng Qiu. 2023.
\newblock \href {https://doi.org/10.18653/v1/2023.findings-emnlp.1055} {{S}peech{GPT}: Empowering large language models with intrinsic cross-modal conversational abilities}.
\newblock In \emph{Findings of the Association for Computational Linguistics: EMNLP 2023}, pages 15757--15773, Singapore. Association for Computational Linguistics.

\bibitem[{Zhang et~al.(2024)Zhang, Zhang, Zhan, Li, Zhou, and Qiu}]{zhang2024speechgptgen}
Dong Zhang, Xin Zhang, Jun Zhan, Shimin Li, Yaqian Zhou, and Xipeng Qiu. 2024.
\newblock \href {https://arxiv.org/abs/2401.13527} {{SpeechGPT-Gen}: Scaling chain-of-information speech generation}.
\newblock \emph{Computing Research Repository}, arXiv:2401.13527.

\end{thebibliography}
